\documentclass[conference]{IEEEtran}
\IEEEoverridecommandlockouts
\usepackage{cite}
\usepackage{amsmath,amssymb,amsfonts}
\usepackage{algorithmic}
\usepackage{graphicx}
\usepackage{textcomp}
\usepackage{xcolor}
\usepackage{titlesec}
\usepackage{listings}
\usepackage{hyperref}
\usepackage{caption}
\usepackage{subcaption}
\DeclareCaptionType{Prompt}
\usepackage{multirow}
\usepackage{cleveref}
\usepackage{framed}
\usepackage{float}
\usepackage{siunitx}
\usepackage{tabto}
\usepackage{booktabs}

\crefformat{footnote}{#2\footnotemark[#1]#3}

\begin{document}

\newcommand{\ra}[1]{\renewcommand{\arraystretch}{#1}}

\newcommand{\code}[1]{\texttt{#1}}

\makeatletter
    \newcommand{\linebreakand}{%
      \end{@IEEEauthorhalign}
      \hfill\mbox{}\par
      \mbox{}\hfill\begin{@IEEEauthorhalign}
    }
\makeatother

\title{ChatGPT4PCG Competition: Character-like Level Generation for Science Birds}

\author{
    \IEEEauthorblockN{Pittawat Taveekitworachai\IEEEauthorrefmark{1}, Febri Abdullah\IEEEauthorrefmark{2}, Mury F. Dewantoro\IEEEauthorrefmark{3}}
    \linebreakand
    \IEEEauthorblockN{Ruck Thawonmas\IEEEauthorrefmark{4}, Julian Togelius\IEEEauthorrefmark{5}, and Jochen Renz\IEEEauthorrefmark{6}}
    \vspace{0.07cm}
    \linebreakand
    \IEEEauthorblockA{
        \IEEEauthorrefmark{1}\IEEEauthorrefmark{2}\IEEEauthorrefmark{3}\textit{Graduate School of Information Science and Engineering}, \textit{Ritsumeikan University}, Kusatsu, Japan\\
    }
    \linebreakand
    \IEEEauthorblockA{
        \IEEEauthorrefmark{4}\textit{College of Information Science and Engineering}, \textit{Ritsumeikan University}, Kusatsu, Japan
    }
    \linebreakand
    \IEEEauthorblockA{
        \IEEEauthorrefmark{5}\textit{NYU Tandon School of Engineering}, \textit{New York University}, New York, USA
    }
    \linebreakand
    \IEEEauthorblockA{
        \IEEEauthorrefmark{6}\textit{School of Computing}, \textit{The Australian National University}, Canberra, Australia
    }
    \linebreakand
    \IEEEauthorblockA{
    \{\IEEEauthorrefmark{1}gr0609fv, \IEEEauthorrefmark{2}gr0397fs, \IEEEauthorrefmark{3}gr0450xi\}@ed.ritsumei.ac.jp, \IEEEauthorrefmark{4}ruck@is.ritsumei.ac.jp, \IEEEauthorrefmark{5}julian@togelius.com, \IEEEauthorrefmark{6}jochen.renz@anu.edu.au
    }
}

\IEEEoverridecommandlockouts

\IEEEpubid{\makebox[\columnwidth]{979-8-3503-2277-4/23/\$31.00~\copyright2023 IEEE\hfill}\hspace{\columnsep}\makebox[\columnwidth]{ }}

\maketitle

\IEEEpubidadjcol

\begin{abstract}
	This paper presents the first ChatGPT4PCG Competition at the 2023 IEEE Conference on Games. The objective of this competition is for participants to create effective prompts for ChatGPT--enabling it to generate Science Birds levels with high stability and character-like qualities--fully using their creativity as well as prompt engineering skills. ChatGPT is a conversational agent developed by OpenAI. Science Birds is selected as the competition platform because designing an Angry Birds-like level is not a trivial task due to the in-game gravity; the quality of the levels is determined by their stability. To lower the entry barrier to the competition, we limit the task to the generation of capitalized English alphabetical characters. We also allow only a single prompt to be used for generating all the characters. Here, the quality of the generated levels is determined by their stability and similarity to the given characters. A sample prompt is provided to participants for their reference. An experiment is conducted to determine the effectiveness of several modified versions of this sample prompt on level stability and similarity by testing them on several characters. To the best of our knowledge, we believe that ChatGPT4PCG is the first competition of its kind and hope to inspire enthusiasm for prompt engineering in procedural content generation.
\end{abstract}


\section{Introduction}

ChatGPT\cite{openai_2022} is software that reached 100 million users within two months of its release, making it the fastest application ever to achieve that milestone\cite{hu_2023}. It is powered by GPT-3.5, a relatively new version of a large language model (LLM) called Generative Pre-trained Transformer (GPT) developed by OpenAI. ChatGPT is a supervised fine-tuned version of GPT-3.5 using reinforcement learning from human feedback to follow human instructions better\cite{ouyang2022training}. Recent LLMs, including GPT-3.5, are remarkable in size, with GPT-3.5 alone having 175B parameters\cite{ouyang2022training}. Other LLMs \cite{rae2022scaling,smith2022using,chowdhery2022palm}, with parameter counts ranging from 280B to 540B, have also been introduced in recent years, enabling unprecedented capabilities not found in smaller models. Wei et al.\cite{wei2022emergent} discovered that once LLMs reach a certain threshold of parameter count, they can perform tasks without explicit training, exhibiting emergent abilities. These emergent abilities have led to the use of ChatGPT in various use cases\cite{info:doi/10.2196/45312, Naumova2023, DOWLING2023103662}.

One notable application of LLMs is in robotics. In a report by Vemprala et al.\cite{vemprala2023chatgpt}, the authors explore different prompt structures to instruct ChatGPT, a conversational LLM, to output commands for controlling robots in various applications. Procedural content generation (PCG) is another area where LLMs could be useful, especially in level generation. In fact, there already exist studies on applying LLMs for PCG. In Todd et al.\cite{todd2023level}, they fine tuned variants of GPT-2 to generate levels for Sokoban, a puzzle game, using only the starting token as an input to the model. Similarly, a study by Sudhakaran et al.\cite{sudhakaran2023mariogpt} also applied a fine-tuned version of GPT-2 to generate Mario levels based on instructions. The key difference between these two recent studies is the former focused on exploring the capabilities of LLMs for level generation, but the later focused on controllable level generation using text prompts.

Inspired by these studies, we want to investigate the application of ChatGPT to PCG further. For this purpose, we have chosen Science Birds\cite{ferreira_2014_a}, a clone of Angry Birds, as our evaluation platform because generating levels with good weight distribution that can withstand in-game gravity is challenging. To make it even more challenging, the generated levels must resemble the shape of an English capital letter. To achieve such a level through ChatGPT, one must carefully and creatively design their prompt.

A prompt, in this context, is a text used to communicate with a language model (LM) to get the desired response. The empirical process of designing a prompt to control the behavior of LMs is prompt engineering (PE)\cite{10.1145/3560815}. Due to its empirical nature, we have decided to host a competition. The objective of this competition is to find the best prompt that can generate a stable level that looks similar to an English uppercase character. Participants are limited to submitting only a single prompt. Each prompt will be evaluated for all 26 English capital letters. We hope that this competition will encourage people to use both creativity and skills in crafting the prompt that would yield the best results. To do so, we enforce some restrictions for simplicity and provide various tools to aid in and accelerate designing the prompt for the competition.

We hope that this competition will push the boundaries of PE and PCG, in particular, level generation. Through the competition, we anticipate that a better understanding is achieved of PE and PCG with LMs. We also hope to spark an interest in not only interacting but also gaining more understanding of ChatGPT and its emergent abilities. Our contributions are as follows:

\begin{enumerate}
	\item We provide tools and a platform, a modified version of Science Birds, for testing ChatGPT responses in generating levels for Science Birds.
	\item We provide a sample prompt that serves as an initial point to modify for better-performing prompts.
	\item We conduct experiments and perform an ablation study on a modified version of our sample prompt to gain a better understanding of its effectiveness and some of the residing components.
	\item We hope to spark an interest in using ChatGPT for PCG.
\end{enumerate}

We discuss related work in Section \ref{sec:related_work}, competition details in Section \ref{sec:competition}, our sample prompt, its modified version as well as variations, and their performance in Section \ref{sec:sample}, and conclusions in Section \ref{sec:conclusion}.

\section{Related Work}\label{sec:related_work}

\subsection{Prompt Engineering}

The alignment techniques in InstructGPT\cite{ouyang2022training} enable LLMs to produce outputs aligned with users' intentions. This leads to the rise of PE in recent years, with various patterns and techniques being developed to improve the performance of LMs\cite{white2023prompt,Saravia_Prompt_Engineering_Guide_2022}. White et al.\cite{white2023prompt} identified six categories of prompt patterns, which serve as reusable templates for achieving specific tasks, such as input semantics and output customization. Participants in our ChatGPT4PCG Competition may find these patterns useful for designing or improving their prompts, or they may simply serve as a catalog of available options.

In addition to prompt patterns, there are several notable prompting techniques that have been developed to improve LM performance\cite{brown2020language,min2022rethinking,wei2023chainofthought,kojima2023large,zhou2023large}. Few-shot prompting\cite{brown2020language, min2022rethinking} provides a few examples of the task in the prompt, allowing for in-context learning and improved performance. Chain-of-thought (CoT) prompting\cite{wei2023chainofthought} guides the model through a reasoning process with a few examples before reaching a conclusion. Zero-shot CoT prompting\cite{kojima2023large} does the same thing, but without any examples. 

Automatic prompt engineer\cite{zhou2023large} is an idea where a scoring model automatically generates and selects the best performing prompt. There are more prompt patterns and prompting techniques that have not been mentioned here, and we encourage participants to experiment and explore some of these techniques feasible for use in the competition, which may result in better performance.

\subsection{Procedural Content Generation}
PCG is any automated or semi-automated technique that aims to generate game content using computer software\cite{shaker2016procedural}. This content encompasses various objects within a game, such as maps, levels, characters, stories, music, and animation. PCG has a rich history and has been extensively utilized to enhance gameplay mechanics and facilitate game designers and artists in the game design and development process. Notably, a particular aspect of PCG is the level generation, which involves creating levels that meet specific requirements, including visual aesthetics and difficulty levels.

In PCG for level generation, quality and diversity are two critical elements\cite{gravina2019procedural}. Quality is the measure of how well the generated levels meet specific criteria, such as playability, difficulty, or playing style. Among these criteria, playability is the most crucial and frequently employed to evaluate the quality of generated levels. On the other hand, diversity determines how different the generated levels are from each other. Insufficient diversity results in levels that are too similar to one another and are often perceived as tedious by players. In contrast, inadequate quality may lead to unplayable levels, which can severely impact the overall gameplay experience.

There already exist studies that use GPT models to generate levels. A study by Todd et al.\cite{todd2023level} found in their preliminary results that GPT-3, which has more parameters than GPT-2, requires less data to be fine-tuned. This suggests that the massive training data used in later versions of LLMs makes them more generic for tasks such as PCG. In the context of LLMs for PCG, our work aims to explore the potential of using ChatGPT in PCG as a standalone approach. ChatGPT is capable of generating coherent and contextually relevant text. By employing ChatGPT for game level generation, we aim to contribute to the growing body of research on using natural language for PCG. In this work, we are keen to examine the efficacy of ChatGPT in producing high-quality game levels.

\subsection{Science Birds}
Science Birds\footnote{\label{foot:science_birds}\href{https://github.com/lucasnfe/science-birds}{https://github.com/lucasnfe/science-birds}} is a platform developed by Ferreira and Toledo\cite{ferreira_2014_a} as an extension to the popular game, Angry Birds, with the aim of fostering experimentation in this genre. The platform is developed on Unity game engine. The level's objects all follow the standardized physics rules implemented by the game's engine. Birds can also come in a wide variety, just like blocks that represent the numerous game objects of varied shapes and materials that can be freely used to compose a level structure.

Science Birds has been widely utilized as a test bed for various research endeavors, particularly in the area of level generation\cite{matthew2017gen, Gam2021, tanabe2021Latent, ebin2020generating, 8080429}. One of them generates varied, stable, and solvable levels composed of multiple self-contained structures\cite{matthew2017gen}. The term ``deceptive games'' was employed to identify an automated method for creating Angry Birds game levels that are deceptive enough to fool modern AI agents. However, these AI agents may contain pitfalls themselves, making them more vulnerable to deception\cite{Gam2021}.

To create game levels with more stability and variety, a technique called deep generative model-based level generation was used in one study\cite{tanabe2021Latent}. This involves encoding the level sequentially as text data, which allows for control of level features through latent variable evolution. In another study\cite{ebin2019angry}, a smile interface system was introduced to Science Birds that features Rube Goldberg Machine mechanisms with a domino effect. The system aimed to promote emotions in spectators while they watched the gameplay.

Jiang et al.\cite{8080429} used pattern-struct and preset-model approaches to create an Angry Birds-like level that resembled a funny quote and won the Fun Track of the CIG 2016 Level Generation Competition. Their approach employed pre-designed sets of blocks that were arranged to form English characters and construct a whole level. This work is similar to ours in generating character-like levels. However, instead of utilizing a pre-defined set of blocks, we challenge participants in our competition to develop general prompts for ChatGPT to generate character-like levels from scratch. 

To evaluate generated levels of Science Birds, several evaluation metrics were used in previous studies. The frequently-used evaluation metrics are stability, block frequency, linearity, difficulty, and density\cite{matthew2017gen, ebin2019angry}. Some studies proposed their PCG approaches in solving specific goals and introduced their own evaluation metrics\cite{kaidan2016procedural, ebin2019angry}. For instance, the study by Abdullah et al. \cite{ebin2019angry} aims to generate levels that feature Rube Goldberg Machine mechanisms, and \textit{dynamic} was introduced to measure objects' movements that indicate the existence of domino effect in the level of interest. However, the most common evaluation metric is stability\cite{matthew2017gen, ebin2019angry, tanabe2021Latent}; stability is important because it closely defines the playability of the level. Hence, we consider stability to be one of our evaluation metrics in this study.

This study also makes use of Science Birds as a platform to visualize and evaluate the efficacy of ChatGPT-generated levels. Specifically, we are interested in evaluating the quality of levels generated by ChatGPT. By leveraging the flexibility and versatility of Science Birds, we aim to gain insights into the effectiveness of this approach and to advance the state of the art in level generation research.

\section{ChatGPT4PCG Competition}\label{sec:competition}

To simplify the evaluation process in this competition\footnote{\href{https://chatgpt4pcg.github.io}{https://chatgpt4pcg.github.io}}, we introduce constraints on the prompt, requiring participants to develop their prompts that result in ChatGPT generating calls to a specific function named \lstinline{ab_drop()}, an example of which is shown in Supplementary Material. We also design metrics accordingly to reflect the stability and similarity of each level, which is the main objective of this competition. Restrictions and rules regarding the prompt and extensive details on the \lstinline{ab_drop()} function are discussed in Section \ref{subsec:prompt_rules}. The evaluation process and metrics on stability and similarity are explained in detail in Section \ref{subsec:comp_eval}.

\subsection{Prompt Rules and Guidelines}\label{subsec:prompt_rules}

In this section, we provide a comprehensive definition and details of the \lstinline{ab_drop()} function, including expected function behavior, accepted parameters, and the required settings for successful execution of the function. Additionally, we discuss prompt rules on various aspects, such as permissible characters and the maximum number of characters. These rules are intended to ensure fairness and efficiency in the competition.

The \lstinline{ab_drop()} function is designed to vertically drop a block from the top and center it at a specific slot, denoted by \lstinline{x_position}. This function shares similarities with the popular game Tetris\textregistered. By defining the function in this manner, we aim to reduce the chance of having overlapping blocks in the level. The behavior of the function is simpler than that of placing a block by specifying x and y coordinates in the level area, which could result in the aforementioned issue. It is worth noting that the \lstinline{ab_drop()} function is functional in the following settings:

\begin{itemize}
	\item A structure is situated on a 2D grid with a width ($W$) of 20 columns and a height ($H$) of 16 rows. The grid consists of 320 cells, each of equal size. The decision to use a grid of this size is based on the consideration that a larger grid could produce more excessively long output from ChatGPT, consuming a lot of tokens, while a smaller grid may not provide sufficient space for ChatGPT to exercise creativity. Therefore, the chosen size is expected to strike a balance between these two factors.
	\item Coordinates $(x, y)$ are used to represent the positions in the grid, where $x$ and $y$ show the horizontal and vertical indices of cells, respectively. For example, $(0, 0)$ denotes the bottom-left corner cell of the grid, and $(W-1, H-1)$ is the top-right corner cell.
	\item A cell on the grid has a size of 1x1. Each cell has unique $(x, y)$ coordinates associated with it.
\end{itemize}

This function accepts two parameters, namely, \lstinline{block_type} and \lstinline{x_position} with the following details:

\begin{itemize}
	\item \lstinline{block_type}: a value that indicates the type of block to be placed. The possible values are \lstinline{b11}, \lstinline{b13}, and \lstinline{b31}, defined in the following and shown in Fig. \ref{fig:blocks}.
		\begin{itemize}
			\item \lstinline{b11} denotes a square block whose size is 1x1.
			\item \lstinline{b13} denotes a column block whose size is 1x3.
			\item \lstinline{b31} denotes a row block whose size is 3x1.
		\end{itemize}
    This limited selection of block types simplifies the output of ChatGPT and the evaluation process.
	\item \lstinline{x_position}: a horizontal index of a grid cell, where \lstinline{0} represents the leftmost column of the grid, and \lstinline{W-1} represents the rightmost column of the grid. The \lstinline{x_position} parameter indicates the center pivot point of the block being placed. For example, if b31 is the only block in the level and is placed at \lstinline{x_position = 4}, it will occupy cells \lstinline{(3, 0)}, \lstinline{(4, 0)}, and \lstinline{(5, 0)}. An invalid position, like a position where a block of interest intrudes on the grid boundary, will result in an error.
\end{itemize}

\begin{figure}[tbp]
    \centerline{\includegraphics[width=0.25\linewidth]{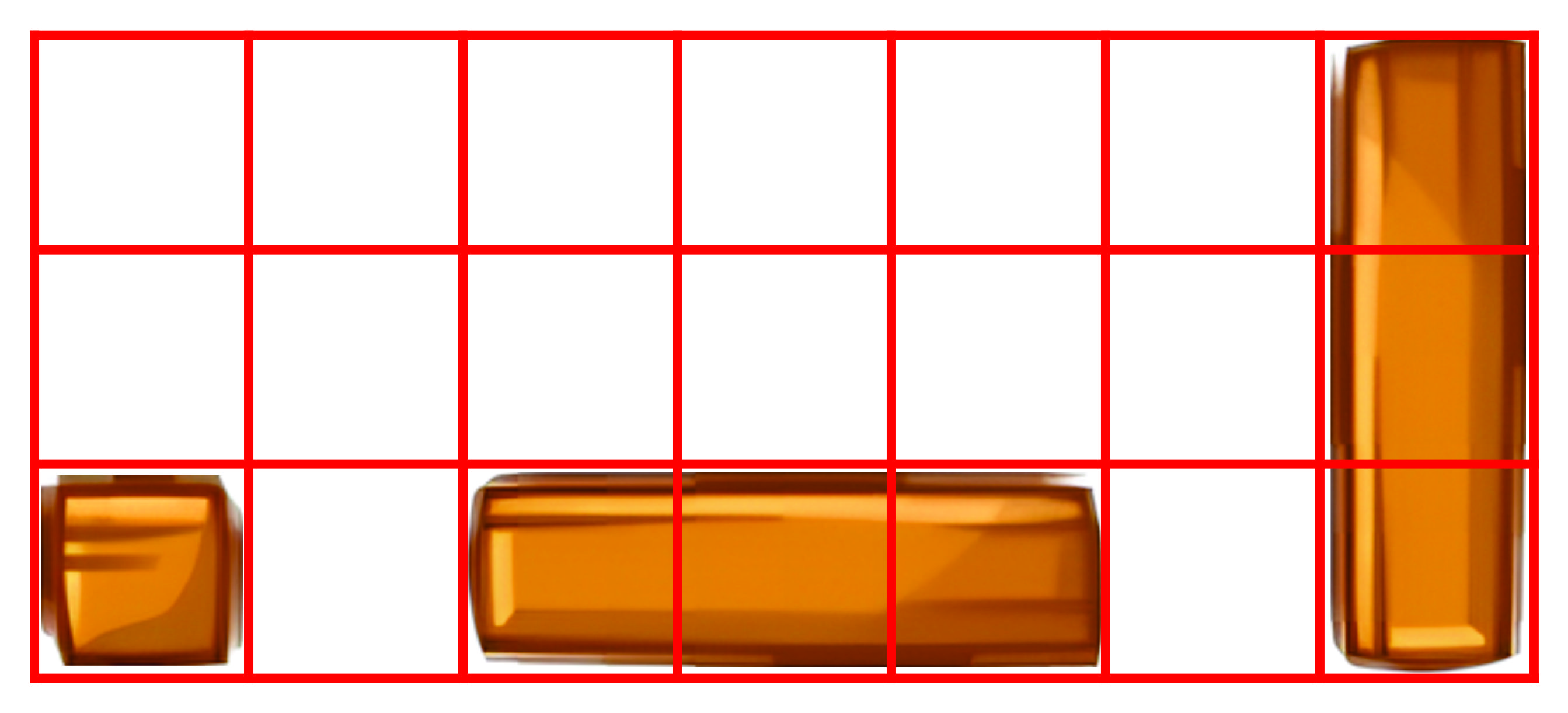}}
    \caption{Block types used in the competition. Starting from the left are \lstinline{b11}, \lstinline{b31}, and \lstinline{b13}, respectively.}
    \label{fig:blocks}
\end{figure}

Next, we introduce several restrictions on the prompt, which consist of the following:

\begin{enumerate}
	\item Prompts must be written in English using only alphanumeric characters, space, and the following symbols: $\sim$ \lstinline{/} \lstinline{\} \lstinline{+} \lstinline{-} \lstinline{*} \textasciigrave~\textquotesingle~ \textquotesingle\textquotesingle~“ ” ‘ ’ \lstinline{.} \lstinline{:} \lstinline{;} \lstinline{?} -- \lstinline{,} \lstinline{!} \lstinline{@} \lstinline{#} \$ \% \lstinline{^} \lstinline{&} \lstinline{(} \lstinline{)} \lstinline{_} \lstinline{=} \lstinline{[]} \lstinline{]} $\{$ $\}$ \lstinline{|} \lstinline{<} \lstinline{>}. This restriction is to ensure that we can better understand them when we select the winning prompt. Furthermore, limiting the prompt to a specific set of allowed characters and symbols helps ensure fairness. This way, no participant can use languages or symbols that others may not have access to.
	\item The maximum word count for a prompt is 900 words. We arrive at this number heuristically based on the maximum number of tokens accepted by the ChatGPT API, which is 4,096 tokens\cite{openai_gpt35} or approximately 1,800 words. This word limit is important because it ensures that all participants have an equal opportunity to create a prompt without exceeding the maximum number of tokens allowed.
	\item Prompts designed for this competition must be created for a one-round conversation. This means that each prompt's interaction with ChatGPT should consist of one user request to and one response from ChatGPT. To maintain consistency in our evaluation process, we disallow prompting techniques that require n-round dialogue. This decision is made because LLMs are stochastic by nature, and we cannot guarantee that we will always be able to use all of the prompts provided by each of the participants--who use such techniques--in succession because the maximum number of tokens may be reached before all of them can be used.
	\item The prompt must include \lstinline{<OBJECT>} to indicate a section of the prompt that will be replaced by us with each target character, such as \lstinline{``I''}, \lstinline{``L''}, or \lstinline{``U''}. Prompts without \lstinline{<OBJECT>} will not be assessed.
	\item To ensure that code blocks can be extracted successfully from responses generated by the ChatGPT API, each output must include three backticks (\textasciigrave\textasciigrave\textasciigrave). All content between the last pairs of triple backticks will be extracted to produce a result file containing a sequence of \lstinline{ab drop()} functions, defined below. It is important to note that the use of loops will be ignored by the script, resulting in only one instance of the loop content being extracted. The use of high-level function approach is chosen based on a study by Vemprala et al. that demonstrated the promising results of generating a sequence of high-level function calls in various applications.
    \item The use of ChatGPT plugins\footnote{\href{https://openai.com/blog/chatgpt-plugins}{https://openai.com/blog/chatgpt-plugins}} is not supported, i.e., all plugins are disabled during the evaluation process. This helps ensure fairness since the plugins only available to limited people, a small number of invited developers and web-based ChatGPT Plus users, and ChatGPT API does not support plugins at the time of writing.
\end{enumerate}
		
\subsection{Competition Evaluation}\label{subsec:comp_eval}

The submitted prompts will undergo an evaluation process that involves subjecting them to 10 trials for each of the 26 uppercase letters of the English alphabet (A-Z). The decision to conduct 10 trials is made to obtain the average performance of each prompt, which may vary over time due to the non-deterministic nature of ChatGPT. The evaluation of the generated levels for each character will be based on their similarity and stability, which are the focal points of this year's competition. The prompts will be scored using the criteria outlined in the scoring policy, which is provided below. To conduct the entire evaluation process, automated scripts and programs are developed and made available at our GitHub repositories\footnote{\label{foot:repos}\href{https://github.com/orgs/chatgpt4pcg/repositories}{https://github.com/orgs/chatgpt4pcg/repositories}}. However, the number of trials and characters in the evaluation set may be adjusted based on the number of participating teams.
		
\subsubsection{Evaluation Environment}
The evaluation will be conducted within a software and hardware environment as specified below.

\begin{itemize}
	\item Software
	\begin{itemize}
		\item OS: Windows 11 Pro
		\item Node.js: 18.15.0 LTS
		\item Unity: 2019.4.40f1
		\item Science Birds Evaluator\footnote{\label{foot:sb_eval}\href{https://github.com/chatgpt4pcg/modified-science-birds}{https://github.com/chatgpt4pcg/modified-science-birds}}, a modified version of Science Birds
		\item Our automation scripts\cref{foot:repos}
	\end{itemize}
	\item Hardware
	\begin{itemize}
		\item CPU: Intel(R) Xeon(R) W-2135 CPU @ 3.70GHz
		\item RAM: 16 GB
	\end{itemize}
\end{itemize}
		
\subsubsection{Evaluation Process}
The evaluation process will be automatically done by scripts as follows:

\begin{enumerate}
    \item The \textit{qualification checking script} will be first run to check for rule violations, including:
    \begin{enumerate}
        \item Whether the prompt contains disallowed characters.
        \item Whether the prompt length exceeds the maximum number of words.
        \item Whether the prompt does not contain the \lstinline{<OBJECT>}.
    \end{enumerate}
    \item The \textit{response gathering script} will load each qualified team's prompt and repeat the following two steps for each character.
    \begin{enumerate}
        \item Replacing \lstinline{<OBJECT>} with the target character
        \item Contacting the ChatGPT API for a specific number of trials. Each trial will always start from scratch and will contain no previous conversation.
    \end{enumerate} 
    \item The \textit{code extraction script} will load each response and produce a new file containing only a series of \lstinline{ab_drop()} commands.
    \item The \textit{text-to-xml conversion script} will load each code file and convert it into a Science Birds level description XML file. Next, \textit{Science Birds Evaluator}, our modified version of Science Birds, will individually load all levels to assess their stability and capture their images. The results of stability will be recorded, and for each level, an image of the structure with black-textured blocks on a white background will be produced by the program. More details about stability evaluation are given in Section \ref{sec:stability}.
    \item The \textit{similarity checking script} will load each image and pass it through an open-source model called vit-base-letter\footnote{\label{foot:vit-base}\href{https://huggingface.co/pittawat/vit-base-letter}{https://huggingface.co/pittawat/vit-base-letter}}. It will then record the similarity result. Details about the model and similarity evaluation are described in Section \ref{sec:similarity}.
    \item Finally, the \textit{scoring and ranking script} will load all stability and similarity results according to the policy defined in Section \ref{sec:scoring} and produce the final rank and score result for all teams according to the evaluation metrics in \ref{subsubsec:eval_metrics}.
\end{enumerate}
		
\subsubsection{Evaluation Metrics}\label{subsubsec:eval_metrics}
Below are our metrics to measure the stability and similarity of generated levels. These metrics serve as evaluation tools and help improve the prompt design. The stability metric evaluates how well the blocks in a level of interest persist after initialization. The similarity metric measures how similar a generated level is to a target character. Using these metrics, prompt designers can improve generation processes for better outcomes. For the rest of the content in this study, $i$, $j$, and $k$ represent the indices of trial, target character, and prompt, respectively.

\paragraph{Stability}\label{sec:stability}
We use \textit{Science Birds Evaluator} to automatically assess stability of levels. We introduce the stability score, $st_{ijk}$, as follows:
$$
st_{ijk} = \frac{total\_blocks_{ijk} - moving\_blocks_{ijk}}{total\_blocks_{ijk}}\text{,}
$$	
where $total\_blocks_{ijk}$ and $moving\_blocks_{ijk}$ are the number of blocks at the initialization step of loading the level into the evaluator and the number of moving blocks during the first 10 seconds after level initialization, respectively. Stable and unstable level examples are shown in Figs. \ref{fig:stable_level_result} and \ref{fig:unstable_level_result}.

This score represents how well a level can withstand the in-game gravity. A threshold of 0.1 is used for the unit of any blocks moved in Unity, which will allow some room for blocks that are moved by quirks in Unity physics and not caused by the structure itself. This metric should encourage participants to create a prompt that can generate a level where most or all of the blocks can withstand the in-game gravity.

\paragraph{Similarity}\label{sec:similarity}
We define the process for evaluating the similarity of a level to the character used in the prompt as the similarity test. The test employs a letter classifier model, described below, and consists of two steps. First, to use the model, an image of the level of interest is required. We utilize \textit{Science Birds Evaluator} to automate the screen capture process. Similarly to the stability test, the screen capture is done 10 seconds after initialization to ensure that the still state of the level of interest is captured--targeting only the remaining blocks and omitting the other details, such as grass and sky. To optimize the capability of the model, the image of each block is replaced by black rectangles, and a white background is used. Captured images always have $1024 \times 1024$ pixel resolution with margins of $100$ pixels on all sides (Figs. \ref{fig:stable_level_black} and \ref{fig:unstable_level_black}).

\begin{figure}[tbp]
     \centering
     \begin{subfigure}[tbp]{0.22\textwidth}
        \centering
        \includegraphics[width=\linewidth]{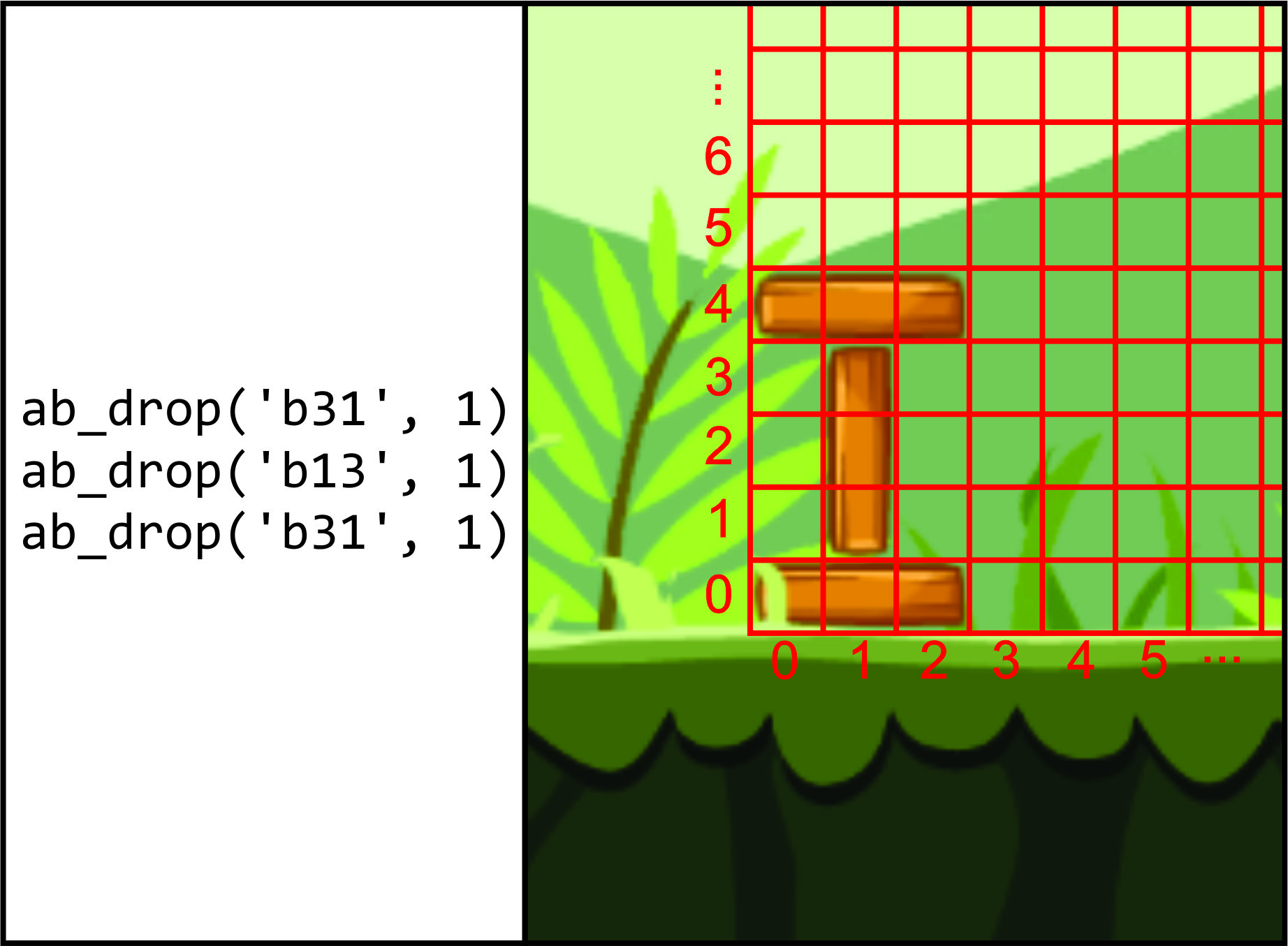}
        \caption{}
        \label{fig:stable_level_result}
    \end{subfigure}
     \hfill
          \begin{subfigure}[tbp]{0.22\textwidth}
         \centering
         \includegraphics[width=\textwidth]{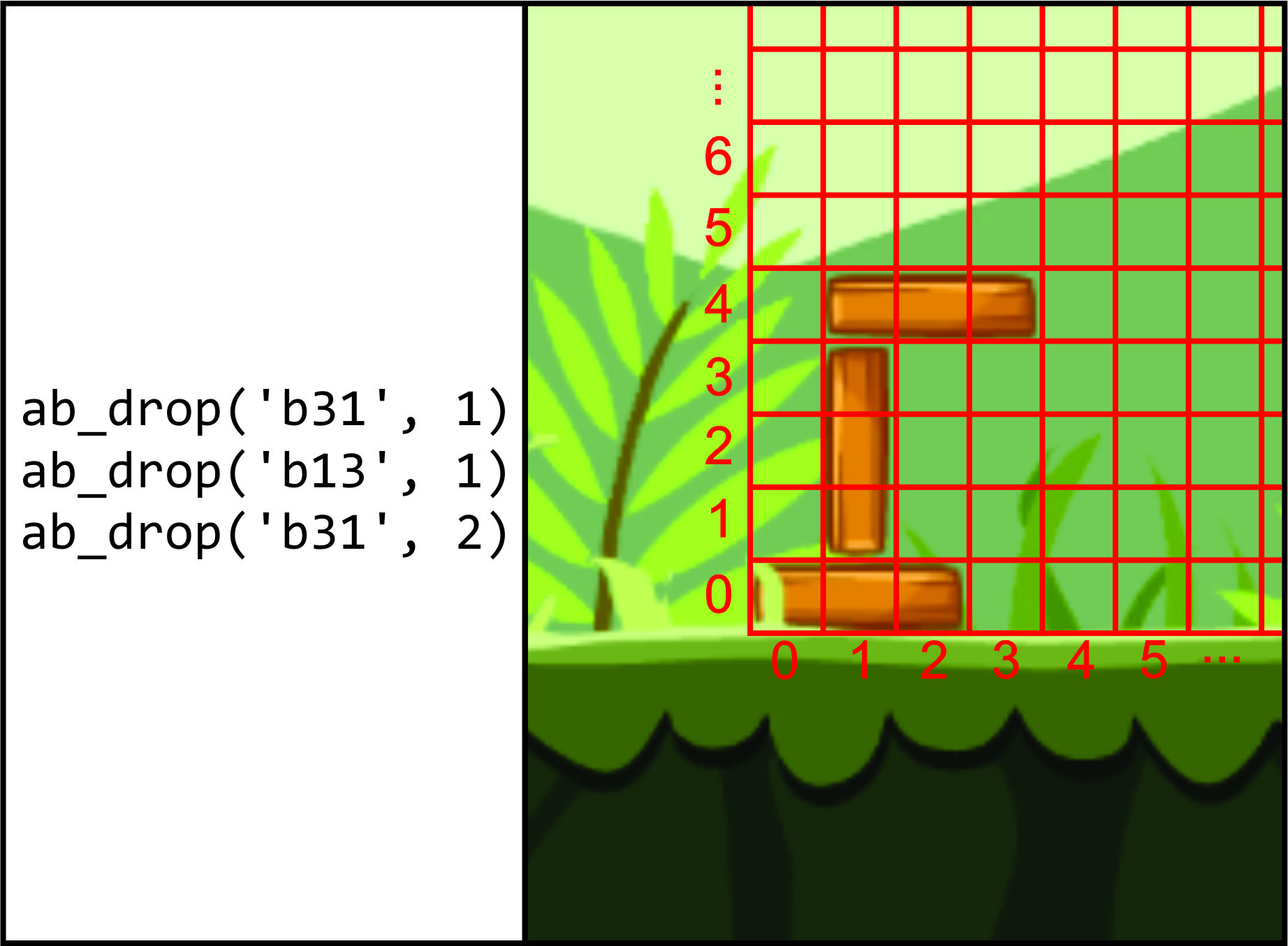}
         \caption{}
         \label{fig:unstable_level_result}
    \end{subfigure}
    \begin{subfigure}[tbp]{0.22\textwidth}
         \centering
         \includegraphics[width=\textwidth]{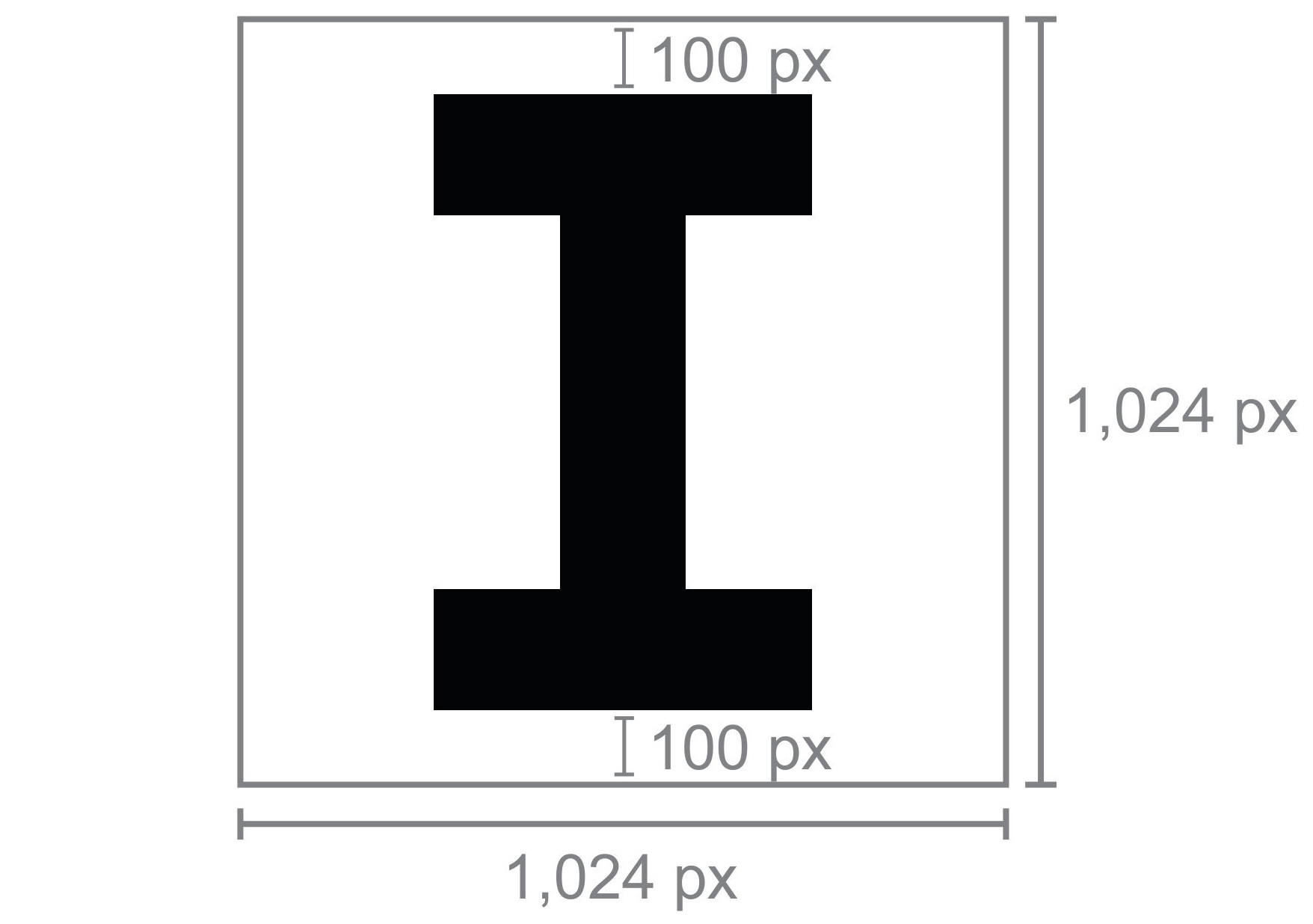}
         \caption{}
         \label{fig:stable_level_black}
     \end{subfigure}
    \begin{subfigure}[tbp]{0.22\textwidth}
         \centering
         \includegraphics[width=\textwidth]{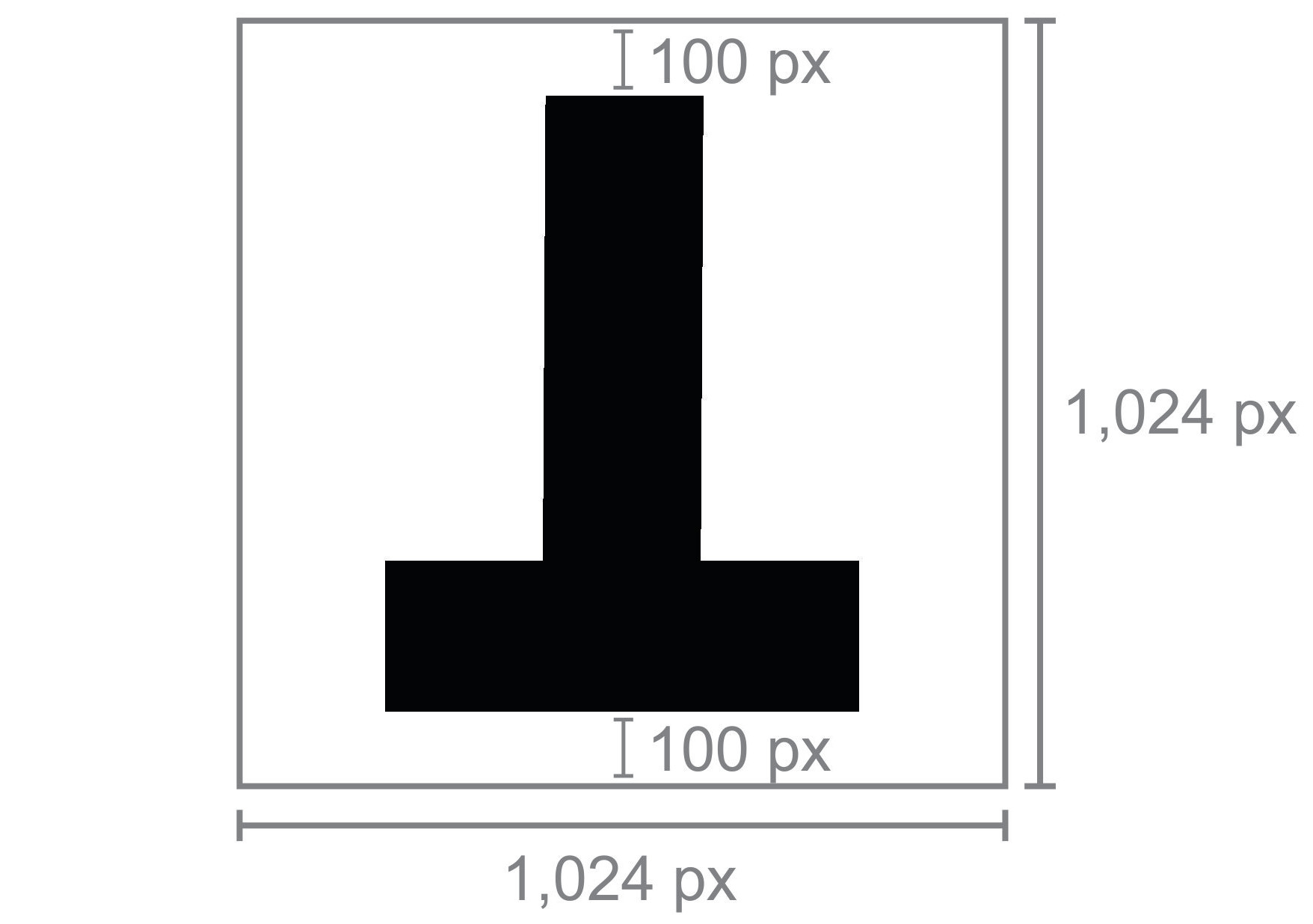}
         \caption{}
         \label{fig:unstable_level_black}
     \end{subfigure}
     \caption{Examples of stable (a) and unstable (b) levels are shown with their respective remaining block images replaced by black rectangles 10 seconds after initialization (c) and (d).}
\end{figure}

The second step is done by inputting the image of a level into the model to calculate the similarity score, $si_{ijk}$, as follows:
$$
si_{ijk} = \sigma (z_{ijk})\text{,}
$$
where $\sigma (z_{ijk})$ is the softmax probability of the model in inferring target character $j$ from the image in trial $i$.

The model is a fine-tuned version of an open-source Vision Transformer (ViT) \cite{dosovitskiy2021image} snapshot, \lstinline{vit-base-patch16-224-in21k}\footnote{\href{https://huggingface.co/google/vit-base-patch16-224-in21k}{https://huggingface.co/google/vit-base-patch16-224-in21k}} fine-tuned and tested on an open-source letter recognition dataset\footnote{\href{https://huggingface.co/datasets/pittawat/letter\_recognition}{https://huggingface.co/datasets/pittawat/letter\_recognition}}. The training set comprises 10,000 images for each class, while the testing set consists of 1,000 images per class. The dataset consists of 26 classes, each representing one of the 26 uppercase English characters that the model is trained to predict. The finetuned model achieved an accuracy of 0.9881 and a loss of 0.0515 on the testing set; hyperparameters and other information are available on the online model webpage\cref{foot:vit-base}. Moreover, using a different model for classification can allow for evaluation in diverse domains.
		
\paragraph{Scoring and Ranking}\label{sec:scoring}
		
Given that $T$, $C$, and $P$ represent the number of trials per target character, the number of characters, and the number of prompts in the competition, respectively, we calculate each prompt's normalized score in the following.

\begin{enumerate}
	\item First, we calculate the weight of each character, $weight_{j}$, to reflect the difficulty of generating a stable and recognizable shape. This approach ensures that characters that are more challenging to generate are assigned a higher weight, while those that are easier to generate receive a lower weight. This weight is set inversely proportional to the normalized average stability and similarity scores, which are calculated as follows:
    $$
        weight_{j} = w\_st_{j} \times w\_si_{j}\text{,}
    $$ 
    where
    $$
        w\_st_{j} = max(1 - \frac{\sum_{k=1}^{P} \sum_{i=1}^{T} st_{ijk}}{PT}, \frac{1}{C})
    $$ 
    $$
        w\_si_{j} = max(1 - \frac{\sum_{k=1}^{P} \sum_{i=1}^{T} si_{ijk}}{PT}, \frac{1}{C})
    $$
    The maximum function, $max$, sets the weight lower bound to ensure that all characters contribute to scoring calculation, at least to a certain extent.
    \item Next, the weighted trial score, $trial_{ijk}$, is calculated to determine the performance or target character $j$ in trial $i$. This score shows how good or bad each trial is when considering both stability and similarity together.
    $$
        trial_{ijk} = weight_{j}st_{ijk}si_{ijk}
    $$
    \item Then, we calculate the average score for character $j$, $char_{jk}$, as follows:
    $$
        char_{jk} = \frac{\sum_{i=1}^{T} trial_{ijk}}{T}
    $$
    \item For prompt $k$, its score, $prompt_{k}$, is calculated by averaging $char_{jk}$ using the following equation.
    $$
        prompt_{k} = \frac{\sum_{j=1}^{C} char_{jk}}{C}
    $$
    \item Finally, the normalized prompt score, $norm\_prompt_{k}$, is calculated in the following. This score will be used for ranking.
    $$
        norm\_prompt_{k} = 100\ \frac{prompt_{k}}{competition}\text{,}
    $$
    where
    $$
        competition = \sum_{k=1}^{P} prompt_{k}
    $$
\end{enumerate}
		
The team that has the highest $norm\_prompt_{k}$ will be declared the winner. If there are multiple teams with the same highest score, the one with the shortest prompt will be chosen as the winner. However, if multiple teams still have the same score and the shortest prompt, they will be considered co-winners.
		
\section{Sample Prompt}\label{sec:sample}

We provide a sample prompt as a starting point for participants to modify and create their own prompts. The use of this prompt is optional, but participants may find the the experiment and results of a modified version of the sample prompt useful in their prompt engineering process. We discuss the structure of the sample prompt in Section \ref{subsec:prompt_structure}. We explain the experiment details of modified prompts in Section \ref{subsec:prompt_exp}. Finally, we present results and discuss their implications of the experiment in Section \ref{subsec:prompt_result}.

\subsection{Prompt Structure}\label{subsec:prompt_structure}

The sample prompt, shown in Fig. \ref{fig:base_prompt}, contains four main sections: (1) instructions, (2) definitions, (3) environments, and (4) tools. The first section provides instructions to ChatGPT on what it needs to do, i.e., ``[...] to generate a stable structure that looks like the \lstinline{<OBJECT>}—the goal.'' which helps guide the output of ChatGPT in a way that it contains a series of \lstinline{ab_drop()} functions. We also provide a marker, i.e., \lstinline{<OBJECT>}, to denote a place where \textit{response gathering script} will replace it with the target character like ``I''. 

Next, we define the construction space where the result of \lstinline{ab_drop()} will occur, giving ChatGPT a general idea about the constraints of the space. Then we define an environment so that ChatGPT knows more about the possible values to the function, in other words, \lstinline{b11}, \lstinline{b31}, and \lstinline{b31} blocks used to construct a level. Finally, we specify, again, the use of \lstinline{ab_drop()} and its definition, so that ChatGPT has more knowledge about how the function works and can use the function to construct the level correctly. These general sections in our prompts are a guideline on what one might include in their prompt. However, we acknowledge that this is not the best possible prompt, and we encourage participants to move, add, or modify the sections to achieve a better outcome.

\begin{figure}[tbp]
\begin{framed}\noindent\small
Use \lstinline{`ab_drop()'} function to generate a stable structure that looks like the \lstinline{<OBJECT>}—the goal. Dropping position and order are crucial.\\

1. Definitions\\
Slots: The map's width is equally partitioned into W slots where W = 20, with slots 0 and 19 being the most left and right, respectively.\\
Layers: The map's height is equally partitioned into H layers where H = 16, with layers 0 and 15 being the bottom and top layers, respectively.
Base: The bottom of the map, i.e., layer 0.\\
\\
2. Environment\\
There are three block types as follows:\\
b11, a square block whose width is 1 unit and height is 1 unit\\
b31, a horizontal block whose width is 3 units and height is 1 unit\\
b13, a vertical block whose width is 1 unit and height is 3 units\\
\\
3. Tool\\
Use the \lstinline{`ab_drop()'} function to vertically drop a block from layer H such that its center is at slot y and drop earlier blocks representing lower parts of the structure.
\end{framed}
\caption{A prompt example provided to our participants.}
\label{fig:base_prompt}
\end{figure}

\subsection{Experiment}\label{subsec:prompt_exp}

The objective here is to assess our modified prompt's performance on some characters to understand its effectiveness. Five variants of the prompt, each with a different set of components, are examined to help us understand the effects of each component on the final outcome. These variants are constructed based on our intuition obtained during the pilot interactions with ChatGPT to develop the sample prompt. The complete prompts for each variant of the modified prompt are available in Supplementary Material. A summary of the variations in each variant, referred to as version (v) from this point, is as follows:

\begin{itemize}
    \item v1: Detailed explanations about stability have been removed
    \item v2: Any code snippets describing the inner workings of the \lstinline{ab_drop()} function have been removed
    \item v3: A mention of ``Tetris\textregistered'' has been removed
    \item v4: Any mentions of the ``block-stacking problem'' have been removed
    \item v5: No components have been removed.
\end{itemize}   

Three types of English capital characters are used: ``I'', ``L'', and ``U''. These characters were selected to obtain a baseline performance for our experiment due to their low level of difficulty in generating a stable level. Each prompt was run 10 times per character, resulting in a total of 150 levels for evaluation.

Although we plan to use the ChatGPT API in the competition, here we ran all prompts using the free version of the web-based ChatGPT Mar 14 Version and utilized ChatGPT Prompt Genius\footnote{\href{https://github.com/benf2004/ChatGPT-Prompt-Genius}{https://github.com/benf2004/ChatGPT-Prompt-Genius}} to export the prompt result to a Markdown file with a \lstinline{.md} extension, one file per trial. If our \textit{code extraction script} failed to extract the response or the \textit{text-to-xml conversion script} failed to convert the level, we re-ran the same prompt using a new conversation session until we obtained a total of 10 trials per character. The remaining evaluation process was conducted using the same process and tools as in the competition, described in Section \ref{subsec:comp_eval}.

\subsection{Results and Discussions}\label{subsec:prompt_result}

The results, assessed using the evaluation metrics in Section \ref{subsubsec:eval_metrics}, are presented in Table \ref{tab:exp_result}. The best version, i.e., the one with the highest $norm\_prompt_{k}$, is v1, followed by v4, v5, v3, and v2. An interesting observation is that even though v4, ranked second, scores the highest for stability in all characters, it does not have the highest $norm\_prompt_{k}$. It is interesting that removing the mention of the ``block-stacking problem'' from the v4 prompt while retaining the instruction about stability appears to result in better stability performance than vice versa. A code snippet is a useful tool for a model to develop knowledge about new tasks, and removing it significantly reduces performance, as observed with the v2 prompt. Additionally, removing the mention of ``Tetris\textregistered'' as in the v3 prompt also negatively affects performance, because the nature of the game, highly likely presented in the training data, might help the model develop intuition.

The v5 prompt, which consists of all prompt components, ranked third among all versions due to not performing well on the \textit{Similarity} for character ``U'', which has the highest $w\_st_{j}$. This is because the evaluation metrics are designed to give more weight to difficult characters, those that a majority of prompts do not perform well in terms of both stability and similarity. It is also suggested that a more detailed prompt may not necessarily yield better performance for this task. For instance, v1 and v4, which are shorter but contain sufficient details, can outperform v5, which is longer.

Table \ref{tab:weights} shows two important observations regarding $weight_{j}$. Firstly, the similarity weights are all equal since all prompts perform well in stability for all characters, as shown in \textit{Stability} in Table \ref{tab:exp_result}. Therefore, the first argument of the max function has a very low value, which is then clipped at the lower bound of $\frac{1}{C}$ for each character. Secondly, \textit{Similarity} plays a major role in the final $norm\_prompt_{k}$ score due to having higher weights for all characters. It can be observed that the character ``U'' has the highest $w\_si_{j}$ and $weight_{j}$, implying that it is the most challenging character to generate a shape that resembles this character, among the three characters in the experiment.

Since the weight of stability is equal, we can focus on only analyzing \textit{Similarity} of version 1, the winner. We found that while it performs moderately well for ``I'' and ``L'', it is the best for ``U''. This demonstrates the capabilities of the metrics used in this competition that encourage participants to design a prompt that works well on all characters for both stability and similarity, in particular those that are challenging.

\begin{table*}[tbp]
\caption{Experiment results where \textit{Stability} and \textit{Similarity} of each variant $k$ denote the average of $st_{ijk}$ and $si_{ijk}$, respectively, among 10 trials for each character.}
\centering
\ra{1.3}
\begin{tabular}{lccccccccccccccccccc}
\hline
 &
  \multicolumn{3}{c}{v1} &
   &
  \multicolumn{3}{c}{v2} &
   &
  \multicolumn{3}{c}{v3} &
   &
  \multicolumn{3}{c}{v4} &
   &
  \multicolumn{3}{c}{v5} \\
  \cmidrule{2-4} \cmidrule{6-8} \cmidrule{10-12} \cmidrule{14-16}\cmidrule{18-20}
  
           & I    & L    & U    &  & I    & L    & U    &  & I    & L    & U    &  & I    & L    & U    &  & I    & L    & U    \\
\hline
\textit{Stability}  & \textbf{1.00} & 0.84 & 0.90 &  & 0.96 & 0.80 & \textbf{1.00} &  & 0.97 & 0.75 & 0.94 &  & \textbf{1.00} & \textbf{0.93} & \textbf{1.00} &  & 0.96 & 0.90 & \textbf{1.00} \\
\textit{Similarity} & 0.18 & 0.53 & \textbf{0.22} &  & 0.22 & 0.24 & 0.02 &  & 0.03 & \textbf{0.65} & 0.01 &  & 0.17 & 0.60 & 0.03 &  & \textbf{0.28} & 0.43 & 0.01  \\
\hline
$norm\_prompt_{k}$ &
  \multicolumn{3}{c}{ \textbf{30.40}} &
   &
  \multicolumn{3}{c}{13.26} &
   &
  \multicolumn{3}{c}{13.36} &
   &
  \multicolumn{3}{c}{23.08} &
   &
  \multicolumn{3}{c}{19.90}\\
\hline
\end{tabular}
\label{tab:exp_result}
\end{table*}

\begin{table}[tbp]
\caption{$weight_{j}$ of each character in the experiment.}
\centering
\begin{tabular}{llll}
\hline
           & I     & L     & U     \\
\hline
$w\_st_{j}$  & 0.333 & 0.333 & 0.333 \\
$w\_si_{j}$ & 0.823 & 0.510 & \textbf{0.944} \\
\hline
$weight_{j}$    &  0.274 & 0.170 & \textbf{0.315} \\
\hline
\end{tabular}
\label{tab:weights}
\end{table}

\section{Conclusions and Future Work}\label{sec:conclusion}

The ChatGPT4PCG competition has been launched at the 2023 IEEE Conference on Games, aiming to find the best prompt capable of generating both stable and character-like levels from responses of ChatGPT. We provided details on the evaluation process and criteria, a sample prompt, and tools for the competition. In addition, we conducted an experiment to assess the preliminary performance of a modified version, of the sample prompt, and its variants on both stability and similarity using the experimental evaluation set. The results showed that version 1 of our prompt variants had the best performance, which highlights the reason why we designed metrics in a particular way. We hope that this competition--which we believe is the first of its kind to focus on PCG for PE--will spark interest in both areas.

We believe that ChatGPT or similar tools can generate not only character-like levels but also word-like, object-like, and image-like levels. Furthermore, the use of ChatGPT for PCG should not be limited to level generation but can also extend to other types of content. In the future, we plan to report on the results and findings from this competition, as well as organize similar competitions with different settings to further explore other areas and make them more challenging while also increasing their impact.

\bibliographystyle{IEEEtran}
\bibliography{references}

\end{document}